# Prediction of Daytime Hypoglycemic Events Using Continuous Glucose Monitoring Data and Classification Technique

Miyeon Jung
Department of Creative IT Engineering
Pohang Science and Technology University
Pohang, Republic of Korea
hoya0604@postech.ac.kr

You-Bin Lee
Division of Endocrinology and Metabolism,
Department of Medicine, Samsung Medical Center,
Sungkyunkwan University School of Medicine, Seoul
06351, Republic of Korea
youbin.lee@samsung.com

Sang-Man Jin
Division of Endocrinology and Metabolism,
Department of Medicine, Samsung Medical Center,
Sungkyunkwan University School of Medicine, Seoul
06351, Republic of Korea
sangman.jin@samsung.com

Sung-Min Park*
Department of Creative IT Engineering
Pohang Science and Technology University
Pohang, Republic of Korea
sungminpark@postech.ac.kr

*Abstract*— Daytime hypoglycemia should be accurately predicted to achieve normoglycemia and to avoid disastrous situations. Hypoglycemia, an abnormally low blood glucose level, is divided into daytime hypoglycemia and nocturnal hypoglycemia. Many studies of hypoglycemia prevention deal with nocturnal hypoglycemia. In this paper, we propose new predictor variables to predict daytime hypoglycemia using continuous glucose monitoring (CGM) data. We apply classification and regression tree (CART) as a prediction method. The independent variables of our prediction model are the rate of decrease from a peak and absolute level of the BG at the decision point. The evaluation results showed that our model was able to detect almost 80% of hypoglycemic events 15 min in advance, which was higher than the existing methods with similar conditions. The proposed method might achieve a real-time prediction as well as can be embedded into BG monitoring device.

## I. INTRODUCTION

Diabetes is one of the most common chronic diseases in the world, affecting 2.72 million individuals (10% of the population) in the Korea [1] and 29.1 million individuals (9.3% of the population) in the USA with increasing incidence [2]. Diabetes can be the cause of kidney failure, lower-limb amputations, and blindness among adults [2]. Achievement of excellent glycemia is most important task to diabetic patients in both type 1 and type 2 diabetes. Diabetic patients should maintain euglycemic blood glucose (BG) levels while all day and be required the wisdom to avoid hyper- and hypoglycemia [3]. Especially, the patients who treated with an insulin are at risk for developing hypoglycemia. Population-based data indicate that 30– 40% of people with type 1 diabetes experience an average of three episodes of severe hypoglycemia each year; those with insulin-treated type 2 diabetes experience about one episode of that each year. Also, individuals with type 1 diabetes experienced about 43 symptomatic (not only severe) episodes annually; insulin-treated individuals with type 2 diabetes experienced about 16 episodes annually [4]. The symptomatic hypoglycemic episode means that the patients feel the symptoms of shakiness, sweating, hunger, irritability or headache [5].

Hypoglycemia is a significant challenge for a precise insulin therapy [6]. When the patient expect his/her future BG change not correctly and make a wrong decision of an insulin injection, hypoglycemic events may occur. Also, daytime hypoglycemia may result from missed or delayed meals, from insulin replacement regimens, or from glucose utilization due to exercise [7]. Many patients express their difficulty in making a decision of their meals, drinkings and insulin injections [8]. Elderly patients receiving insulin treatments should monitor their BG themselves, but they have few knowledge to manage hypoglycemia. Some patients said they do not use enough amount of insulin because of worry about hypoglycemia. While the symptoms of hypoglycemia alert individuals to an impending episode, these warning signs can diminish quality of life and discourage tight glycemic control [9]. The symptoms elicit anxiety and fear of future hypoglycemia, and they are a practical barrier to safely and effectively achieving their glycemic goals [9]. Sometimes, even physicians are unaware of the multitude of consequence of hypoglycemia or how to deal with them. Daytime hypoglycemia causes serious social disruptions for the patients because it interferes with daily activities such as working, attending school, or driving [4]. Also, if patients cannot detect their hypoglycemia, hypoglycemia unawareness, they can fall into more dangerous situations. Therefore, patients should be supported by a proper technology to predict daytime hypoglycemia in real-time.

Untreated hyperglycemia typically causes logn-term damages such as cardiovascular disease, nerve damage or kidney damage [10]. On the other hand, hypoglycemia accompanies impending effects such as seizure, loss of consciousness or death [9]. Patients with diabetes play with a double-edged sword when it comes to deciding BG and A1c target levels. On the one side, tight control has been shown to be crucial in avoiding long-term complications; on the other hand, tighter control leads to an increased risk of iatrogenic hypoglycemia, which is compounded when hypoglycemia unawareness sets in. Even though both of hypoglycemia and hyperglycemia are undesired outcomes, hypoglycemia may occur with any insulin administration in diabetic patients, and results in dangerous and prompt situations. For example, the insulin treatment can affect a driving condition of the patients and causes car accidents [11]. Hypoglycemia indicating an impaired ability to drive, retinopathy formation impairing the vision needed to operate a motor vehicle, and neuropathy affecting the ability to feel foot petals can impact driving safety. In a recent study, 62% of health care professionals suggested that insulin-treated drivers should test their blood glucose before driving certainly [11]. The most significant subgroup of patients is drivers managing their diabetes with insulin [6].

For the patients who are under insulin treatments, the level of effort required and the difficulty in achieving glycemic targets lead to fatigue and burnout. Insulin treatment is a key driver of hypoglycemia. One large population-based study reported an overall prevalence of 7.1% (type 1 diabetes mellitus) and 7.3% (type 2 diabetes mellitus) in insulin-treated patients, compared with 0.8% in patients with type 2 diabetes treated with an oral sulfonylurea [9]. Use of medical devices such as continuous subcutaneous insulin infusion (CSII, so-called insulin pump) and continous glucose monitoring (CGM) device can improve glucose control and insulin therapy. Also, low-glucose suspend device (LGS) and Predictive low glucose suspend device (PLGS) may offer the potential of hypoglycemia prevention. A Low-glucose suspend device is an insulin pump that shuts off automatically to stop insulin delivery when an integrated CGM sensor detects that the patient's BG has fallen below a threshold (60 - 90 mg/dL) and the patient fails to respond to an initial alarm. However, the Low-glucose suspend device does not always prevent hypoglycemic events, because patients may already be hypoglycemic when the alarm is activated. Recently, Predictive low glucose suspend device (PLGS) has been introduced. It generates alarms, then reduces or stops insulin delivery before BG gets too low [12]. Minimed 640g (Predictive low-glucose suspend) is currently available in Europe and has received FDA approval in the USA. This hypoglycemia minimizer function (Predictive low-BG alarm) is a crucial part of an insulin therapy to ensure the safety of patients.

However, those hypoglycemia minimizers still need additional innovations to be adopted by more patients against practical barriers in real situation. For example, patients who do not use an insulin pump also should be alerted to the possible onset of hypoglycemia. Currently Minimed 640g' algorithm requires inuslin delivery information, so only for the patient who has both an insulin pump and a CGM device – Sensor-augmented pump (SAP). A prediction algorithm embeded in the CGM device should be developed for the patients who use only CGM with a sufficient prediction accuracy.

Our proposed prediction method for the patients having a CGM device is based on the simple observation that future hypoglycemic events are likely when the current value of BG is low and decreasing rapidly. The trend and speed of change help patients assume future BG as a tacit knowledge of general self-management, but our CART model will help patients predict impending hypoglycemic based on the statistical learning of real evidence. Also, we use a peak value after meals as a starting point to calculate the rate of decrease. We try to reflect the impact of a meal on future BG levels when we select predictors, and the pattern of BG decreasing after meals can be reflected well as we use the 'peak' value (using the peak value to calculated the rate is better than a interval velocity). We transformed original BG time series to generate informative features for a prediction model. Use of a decision tree is good to interpret the meaning of classification rules. Decision tree methodology is a commonly used data mining method for establishing classification systems based on multiple covariates or for developing prediction algorithms for a target variable, and it has been used often in previous research to predict the risk of dangerous clinical events [13].

The contribution of this paper is twofold. First, this is the first analytical study for modeling multiple patients' CGM data without insulin information, only focusing on daytime hypoglycemia. Based on the idea that the mechanisms of daytime hypoglycemia and nocturnal hypoglycemia are different, we propose a different prediction model using meal time information. Daytime hypoglycemia is more difficult to be predicted because there are many factors affecting BG during daytime such as meals, snacks, and exercise. Secondly, the proposed method has a clinical impact for enhanced daytime hypoglycemia prevention because it achieves high prediction accuracy and sensitivity. Also, this study shows the clinical possibility of real-time prediction by simplifying the prediction algorithm.

Section II reviews related work. Section III outlines a classification technique used in our prediction model. Section IV describes our data and pre-processing phases. Section V describes our proposed prediction model. Section VI is the conclusion.

II. RELATED WORK

**Hypoglycemia Prevention**
Hypoglycemia prevention research is related to development of an artificial pancreas (so-called closed-loop system) [14]. Artificial pancreas is the most promising beta-cell replacement therapy for the patients with type 1 diabetes incorporating CGM sensors and insulin pumps. The simplest form of the automated closed-loop system is suspension of insulin delivery at a low glucose threshold - low-glucose suspend (LGS) device. A number of clinical and *in-silico* studies attempted to deal with the hypoglycemia prevention - mainly nocturnal hypoglycemia - as the primary control objective [6], [15]–[20]. In [17], a hybrid closed-loop system was employed during night period using model predictive control (MPC) algorithm, and aimed at regulating

BG levels overnight to avoid nocturnal hypoglycemia. Earlier versions of MPC algorithm were tested in previous clinical studies to evaluate their control and prediction performance during overnight fasting conditions [15],[21]. These studies concluded that the MPC algorithm is well suited for glucose control under overnight fasting conditions in type 1 patients. Other studies also reported their results of the LGS and PLGS features reducing nocturnal hypoglycemia significantly. The objective of the study [22] was evaluating the LGS feature of the Paradigm Veo insulin pump (Medtronic, Inc., Northridge, CA). The Paradigm Veo insulin pump can automatically suspend basal insulin delivery in the event of CGM-detected hypoglycemia, thus reducing the duration of hypoglycemia. The study concluded that the LG feature was associated with reduced nocturnal hypoglycemia. On the other hand, 20 episodes (34% of daytime LGS episodes) lasted the maximum 2 hr even after suspension of insulin delivery. The reason why daytime hypoglycemia couldn't be prevented by the LGS feature is that the function only suspended basal insulin delivery, but daytime hypoglycemia is strongly related with bolus insulin. The study [23] aiming prevention of nocturnal hypoglycemia using PLGS also reported impressive results of nocturnal hypoglycemia prevention; Current closed-loop system is superior to the control in preventing overnight hypoglycemia – night-time-only artificial pancreas [14]. Even up to 75% of hypoglycemia seizures occur at night, we thought that new methods should be studied to prevent daytime hypoglycemia well, and decided to focus on daytime hypoglycemia in this study.

Daytime hypoglycemia is under complex conditions because patients keep them active during daytime. Especially, postprandial hypoglycemia is mainly implicated by a bolus insulin (rapid acting insulin). Therefore, the prediction method should be designed to cope with shortly changing BG. There is one study focusing on daytime hypoglycemia [24]. The prediction model of [24] predicted BG levels better than diabetes experts did, and could be used to anticipate almost a quarter of hypoglycemic events (true positive ratio = 23%) using CGM, carbohydrate intakes, and the amount of rapid acting insulin with $L$ = 30 min. However, sensitivity of 23% is not sufficient in real situations, although the authors argued that most of false alarms were in near-hypoglycemic regions. The sensitivity (true positive ratio) of prediction should be improved even with smaller $L$ than this model, and recording the carbohydrate intake and the amount of rapid acting insulin may be burdensome task for the patients. We determined the optimal $L$ having trade-off relation with the accuracy, and tried to use only CGM data for better usability. Other studies didn't focus on the short-term scrupulous avoidance of daytime hypoglycemia. They developed general algorithms regardless of the time of hypoglycemia [25], [26].

Previous studies of hypoglycemia prediction also are lacking in prediction performance. The study [27] using recursive autoregressive partial least squares models predicts hypoglycemia based on the past glucose variations. A sensitivity of 86% and specificity of around 60% are obtained for the algorithm. However, the false alarm rate of 0.42 false positive/day is not sufficient. We thought different alarm systems should be used for two hypoglycemia to get higher accuracy than existing methods.

Also, there are conference proceedings and journal articles aiming hypoglycemia prediction but with small subjects less than 10. Recruiting patients using CGM device may be so difficult, but we believe that around twenty is the minimum number to provide academic generality. Also, most of previous studies focused on type 1 patients only because CGM device usually is for the type 1 patients having a strong difficulty of managing BG. However, many type 2 patients also have big challenges to manage their BG, and some have more severe problem of hypoglycemia than type 1 patients. Fortunately, we could acquire enough sample size of type 2 patients from the medical center, so we plan to develop the hypoglycemia prediction method using the data from both of type 1 and type 2 patients.

For hypoglycemia prediction, there are two major methods, time series estimation and classification techniques. [20], [26]–[30], [33] used original time series itself and predict future BG using regression-based continuous model. On the other hand, [24], [25], [31], [32] used machine learning methods as classification techniques (Table 1). Time series prediction contains various models such as an autoregressive moving-average (ARMA) model, AR model, MA model, ARIMA model, or ARMAX model. The concept of time series models is that the current glucose measurements can be described as a linear function of previous glucose measurements and residual terms. Therefore, this time series method is suitable to the study whose objective is the estimation of future BG values as real number; e.g. for fully-automated closed-loop. We thought the machine learning methods as event-based classification techniques is proper to simply predict future hypoglycemia events. The machine learning methods are robust to instant changes of BG, and don't need additional alarm decision rule rather than time series estimation needs additional alarm decision rule based on the predicted single value.

**Trade-off between Usability and Prediction accuracy**

Based on the literature review, we found that 1) time lag $L$ and 2) required input variables are most important aspects of evaluating a hypoglycemia prediction method. These two quantities are related to both usability and accuracy of prediction, which are important research considerations in hypoglycemia prediction.

Based on our definitions, $L$ is used to mean the shortest guaranteed time between the observation window and the prediction horizon $H$. The observation window is the time between (1) the first report of a datum to a decision algorithm and (2) the moment at which the algorithm decides whether hypoglycemia will or will not occur within defined $H$. For example, if a hypoglycemia alarm was emitted at 10:00 based on past data, and the prediction is that a hypoglycemic event will occur between 10:15 and 10:25, $L$ = 10:15 - 10:00 = 15 min and $H$ = 10:25 - 10:15 = 10 min (the left one of Figure 1).

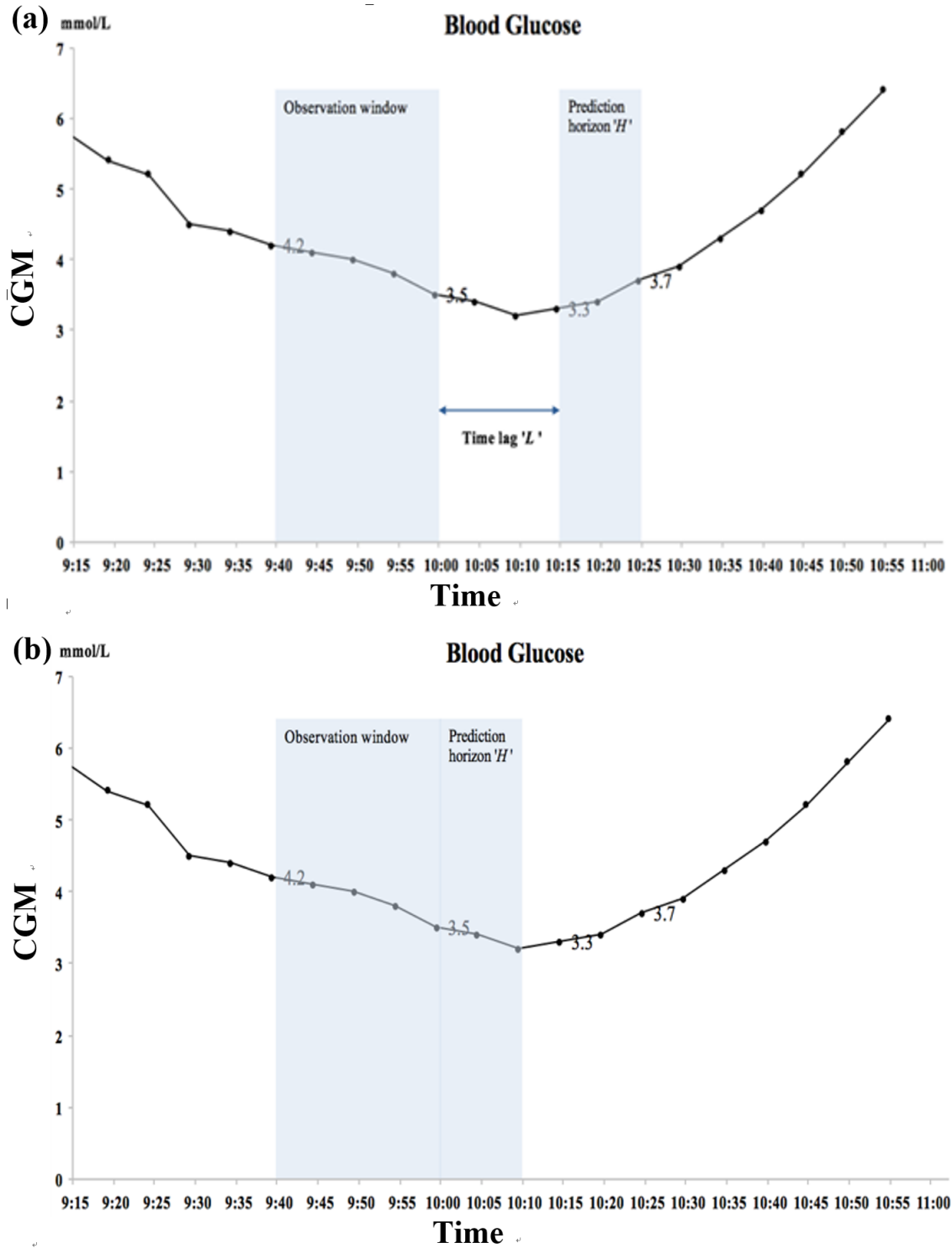

**Figure 1. Our definition of Time Lag '$L$'; Left figure's $L$ is 15 min; Right figure's $L$ is zero.**

$L$ is an important feature of hypoglycemia prediction research. The study [22] proposed a multitier alarm system depending on the length of $L$. Using the 20-min $L$, [22]'s algorithm issues a mild warning, more for information purposes than a call for corrective action. A shorter $L$ of 10 min can be used for a higher level of warning, asking the user to keep close watch. If the 5-min prediction detects hypoglycemia, then the algorithm can issue an alarm requesting corrective action. If $L$ is too long, the prediction algorithm cannot use recent data points because $H$ is far from the decision point. Therefore, as $L$ increases, the accuracy of prediction decrease. On the other hand, if $L$ is too short, the patient may not have time to react to prevent hypoglycemia; i.e., the alarm is useless. The study [22] proved that the sensitivity and specificity of the prediction get low with long $L$, but the mean sensitivity of their method with 15-min $L$ and 70mg/dL threshold (our conditions) is around 40%, lower than our result. Most studies did not have enough $L$, and some had enough $L$ (~15-30 min) but low accuracy. Some authors argued that their model has the $L$ of more than 30-min, but the argument has a large flaw: a prediction of a hypoglycemic event 'within the next 30 min' means that one or more hypoglycemic event occurred during this interval; the weak point of this statement is that most hypoglycemic events within 5-min (very the front part of 30-min interval) is equivalent to all events within the 30-min interval. If many hypoglycemic events occur within 5-min later, the prediction accuracy looks good, but patients are not guaranteed to receive a prediction of a hypoglycemic event that occurs 15-min later.

Second, 'required input variables' are the data that must be acquired to predict hypoglycemia. The required variables depend on the study design. For example, a prediction model that uses both CGM data and meal menu or the amount of insulin can achieve higher accuracy than a model that uses only CGM data. One previous study [33] used heart rate variability (HRV) patterns as an additional predictor with CGM data, and yielded a sensitivity of 79% and specificity of 99%. As a result of literature review, we concluded that if many input variables are used, the accuracy of prediction increases, because the wide range of available information helps researchers increase the precision of their prediction model. We reviewed the required input variables and the $L$ used in previous studies (Table 1).

**Table 1. Representative Prediction Methods in Previous Studies**

| Method | Required input variables | Description & Advantage | Drawbacks | Time lag $L$ | Sensitivity | Specificity |
|---|---|---|---|---|---|---|
| Recursive and AR Time-series Model [28] | CGM | The previous CGM for the previous sampling times are used for linear time-series regression, and the 'recursive' means the algorithm updates the observation window and prediction horizon with a moving window. | Time-series model is heavy algorithm. [28] is for only subjects with T1DM, and needs to balance accuracy and time lag. | 10, 20, 30, 40, 50-min | 86% (30-min) | 58% (30-min) |
| ARMAX Time-series Model [26] | CGM, Insulin on board, and Physical activity | The previous CGM are used for ARMAX time-series regression. The algorithm [26] showed high accuracy and can be used for both offline and online prediction. | Time-series model is heavy algorithm. [26] is for only subjects with T1DM, and requires insulin and activity information. | 30-min | 100% | 100% |
| Logistic Regression based Classification Model [27] | CGM, HRV (heart rate variability) | HRV and CGM are used as the predictors of the logistic regression classifier. | Short time lag, Small sample size of hypoglycemic events (16 events) | 10-min | 79% | 99% |
| Support Vector Regression Model [24] | CGM, Carbohydrate intake, The amount of rapid acting insulin | The state vector computed by meal, insulin and glucose dynamics is used to predict hypoglycemia, and the SVR model was trained on individual patients. | Inferior prediction accuracy, Inconvenient input variables. | 30-min | 23% | 99.2% |
| Machine Learning Model [25], [34] | SMBG([25]), CGM([34]) | Past BG levels are transformed into specific discrete predictors, and these predictors are used as input for machine learning models such as decision trees, SVM, or Naïve Bayes. These models are easy to understand. | Additional calculation time to extract prediction parameters. [34] was too lack of subjects (less than 10). [25] was only for the one-day prediction window. | [34] 30-min | [34] 86.5% | [34] 96% |

In this study, we used CGM data and meal time, because we considered that recording meal menus or the amount of insulin is too burdensome to users even though it provides better accuracy than a simple meal-time notation. This method can be used as a decision support function on CGM device, to alert the user of future daytime hypoglycemia.

III. METHODS

The classification and regression tree (CART) was used to predict risk of hypoglycemia. To the best of our knowledge, only one research used a decision tree method [32]. The CART method is briefly described here.

*A. Classification and Regression Tree (CART)*

CART is one of decision tree methods (CART, J4.8, C4.5, CHAID or REPTree), a recursive partitioning method for classification. CART is developed by Breiman et al. [34] in 1984, the algorithm repeatedly do binary split of each independent variable, building a tree to classify each object into a class. The process of CART is following;

*1) Growing the tree :* The step of building tree using a training set (part of whole data in cross-validation) has severeal decision issues such as selection of independent variables and splitting criterion.

*2) Prunning :* If fully-grown tree is too complicate, the classficiation rule is also noninterpretable. By prunning a part of nodes of fully-grown tree, we can get a simple tree.

*3) Classification :* Using the pruned tree, we can produce a classfication rule, and predict the class of new data or a test set of existing data (for the cross-validation method). We finally yields the output of the CART ($H(t)$ in our study) by following each nodes of if-then rules.

Figure 2 is the tree learned on the training set except $1^{st}$ test group decided by the $1^{st}$ random allocation in our study. The tree shows the rule of classification to one of two classes N and H; N = non-hypoglycemia, and H = hypoglycemia. The splitting criteria are expressed using $x_1$ and $x_2$; The $x_1$ is the rate of decrease, and the unit is (mmol/dL)/min. The positive value of it means decreasing BG. The $x_2$ is the absolute value at a decision point, and the unit is mmol/dL. The characters N and H are terminal nodes, which means a predictive class (output of the tree algorithm). The initial node is called root node. In this tree, the root node means that *an object needs to be moved to the next node when $x_2 < 6.45$ mmol/dL, or an object terminally is classified to N, when $x_2 \geq 6.45$ mmol/dL*. This rule of root node makes sense, because when the absolute value before 15-min is higher than 6.45 mmol/dL, the BG level isn't able to drop until 3.9 mmol/dL (hypoglycemia) within just 15-min. The advantage of decision tree is that researchers can interpret the meaning of each node easily.

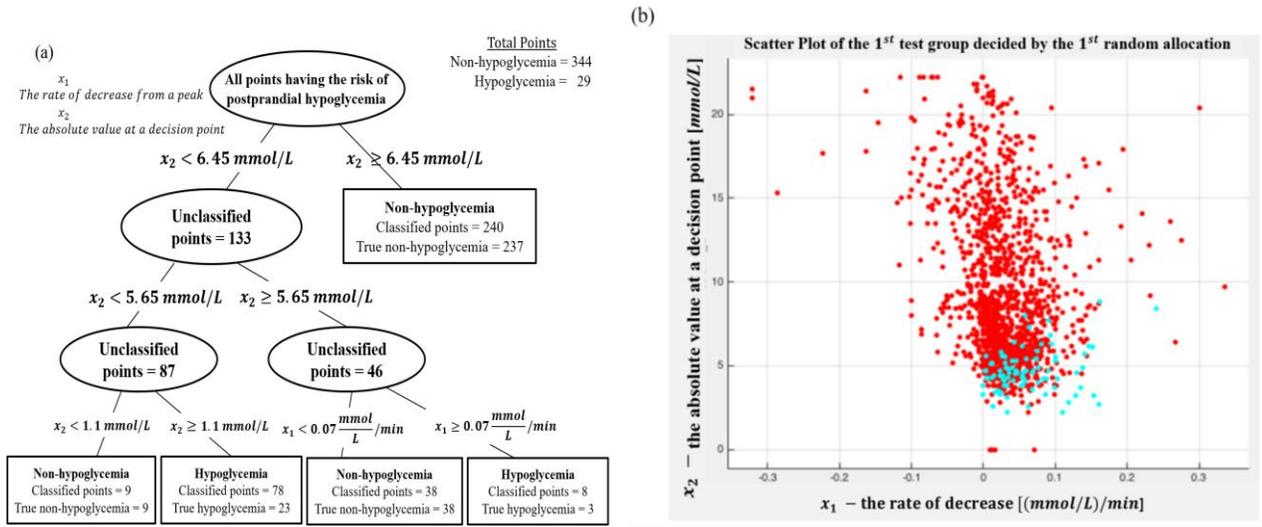

**Figure 2. Example of CART (left) and a scatter plot of training set (right); Red dots mean hypoglycemia, and blue dots mean non-hypoglycemia**

We only used two predictors. Therefore, the data space is 2-dimensional (the right one of Figure 2), and the CART algorithm splits the data space by choosing one predictor (axis) repeatedly. In real situations, this splitting task is imperfect; if it uses too many split lines, all data can be classified correctly, but the tree is strongly adapted (overfit) to the training set, so they do not work well for new data. To solve this problem of too many nodes in a decision tree, cost-complexity measures or expert's knowledge can be used to eliminate (prune) a few nodes. In this study, we used the Gini index as the criterion to reduce the impurity of child nodes at each parent node, and pruned the tree based on expert's knowledge.

*B. Preditction Evaluation Method*

CART was applied to the data from 33 diabetic patients for prediction model learning and evaluation of prediction. The evaluation was based on 5-fold cross-validation between the training set and the test set. Cross-validation is popular method to evaluate classification rules. We divided whole objects (data points) into five groups using a random allocation function in EXCEL, and one group among them was selected as a test set in sequence. Then, a tree learned on the training set (other four groups except the selected test set) decided by the random allocation function. Performance result was yielded when we tested a tree to the corresponding test set. We totally did the random allocation four times. Therefore, totally 20 ($5 \times 4$) performance vectors (accuracy, sensitivity, and specificity) were yielded. The left tree in Figure 2 is the tree learned on the training set except $1^{st}$ test group decided by the $1^{st}$ random allocation. The class was blinded for the test set, so it is like new data in real situations. Our model predicts onset events regardless of patients since the blood glucose variations over time have similar properties in any diabetic patient i.e., we regard each object to be an individual event predicted by the past BG trend. Therefore, the measures were calculated to evaluate prediction of individual events. We also tested the prediction model to individual patients and specific patient groups.

IV. DATA AND PRE-PROCESSING

*A. Data Acquisition*

We collected CGM data from 109 patients, then selected patients who had experienced daytime hypoglycemia more than once during data collection period; each patient included at least two episodes of hypoglycemia between 7:00 a.m. and 11:00 p.m.; the reference of daytime. The patients used a diagnostic CGM. The diagnostic (professional) CGM devices are worn for a fixed time period while the user is blinded to the data. Diagnostic CGM enables retrospective analysis and identification of glucose trends and patterns, and provides a complete picture of blood glucose. Patients who only experienced nocturnal hypoglycemia were not included in the data set for prediction model learning, because we decide to focus on daytime hypoglycemia.

To reduce the dimension of the raw data, we extracted only meal time and BG level at each time stamp. The final data consisted of 33 patients' data sets of ~72 to 96 h of 5-min CGM readings. Among 33 patients, 21 patients have type 1 diabetes, and 10 patients have type 2 diabetes. Other 2 patients are in a dumping syndrome caused by total gastrectomy surgery. Even the two patients are not in diabetes, they also experienced daytime hypoglycemia, and can be potential users of CGM. The few missing BG data were indicated as N/A. Blood samples were drawn for reference BG measurements at every meal during the day, so the meal time is the time at which a reference measurement has been recorded. The meal time is important information to calculate predictor variables and to establish the cohort for this study.

**Table 2. CGM data example of one patient; the forth column is meal time indicator.**

| Sample # | Date | Time | Meal (reference measurement) | Sensor Blood Glucose |
|---|---|---|---|---|
| 0 | 7.Sep.15 | 9:22 | . | 11.8 |
| 1 | 7.Sep.15 | 9:27 | . | 11.4 |
| 2 | 7.Sep.15 | 9:32 | **10.2** | 11.8 |
| 3 | 7.Sep.15 | 9:37 | . | 12.2 |
| … | … | … | … | … |

A binary attribute for hypoglycemia is added to the collected data based on our selected criteria; its value is '1' when BG ≤ 3.9 mmol/L ≈ 70 mg/dL, and '0' otherwise. This threshold was the choice of the physician that cooperated in this study. In a previous literature, hypoglycemia is accompanied by a measured plasma glucose concentration 70 mg/dl (3.9 mmol/l) [35]. Then, we transformed the data to create the predictive features for CART model.

*B. Assumptions for Preprocessing Based on Blood Glucose Dynamics*

We transformed raw data to generate informative features (predictors) for a decision tree based on our assumptions. We confirmed our assumptions of blood glucose dynamics after meals based on literature. We build the model of BG after meals to extract predictor variables.

First assumption is that, blood sugar level reaches to their highest level within two hours after meals. In [36], for this retrospective analysis, BG level reaches to their highest levels within 90-min after meal in 80% of patients, Also, peak times were similar after breakfast, lunch and dinner, and in type 1 and type 2 diabetic patients. Based on this study, we set a range of two hours more conservatively to do not miss peak value. To predict hypoglycemia, we use two previous CGM data points to calculate the rate of decrease in BG (first predictor). In detail, the rate of decrease in BG is the difference between the highest level and current level (at a decision point) divided by time difference between them. Therefore, we need to catch the highest level (peak value) after meals. To achieve the peak value, we set a range of two hours after meal time, and found the highest level using a MAX function in EXCEL.

Second assumption is that there is a possibility that hypoglycemic events may occur from 2hr 15min to 4hr after a meal (postprandial insulin injection); that is, there is little hypoglycemic event within 2 hr 15 min after a meal. Many previous studies reported that insulin-induced hypoglycemic event after a meal occurs minimally after 2 hr [37]. This assumption let us to set rational time range covered by our prediction algorithm. To find a peak value after meals, we should set the starting point of first decision making.

Third assumption is that, to avert hypoglycemia, a patient should be alerted to its possible onset at least 15-min in advance. Previous study said, after simple carbohydrates intake, it takes minimally 15 minutes to increase BG level [38]. Also, glucagon injections can stabilize patients' BG within approximately 15 minutes [39]. As we explained, time lag $L$ is very important research decision. Although the longer is the better for the $L$, previous studies with 30-min $L$ [24], [28] had low sensitivity or specificity (The study [34] reported both high sensitivity and specificity, but the evaluation was conducted on only 6 patients). We conclude that the alarm triggering patients' prompt actions in 15-min advance is a great help to avoid disastrous situations.

## C. Preprocessing to Extract Predictive Features of CART model

Let X = $\{x_t, t=1,…,n\}$ be the time series of CGM data. The time interval between $t_1$ and $t_2$ is 5 min, because the BG level was detected every 5 min by CGM device. The unit of $x_t$ is mmol/L. Y = $\{y_t, t=1,…,n\}$ is the binary hypoglycemic time series corresponding to each X { 1 for a hypoglycemia, 0 for a non-hypoglycemia }.

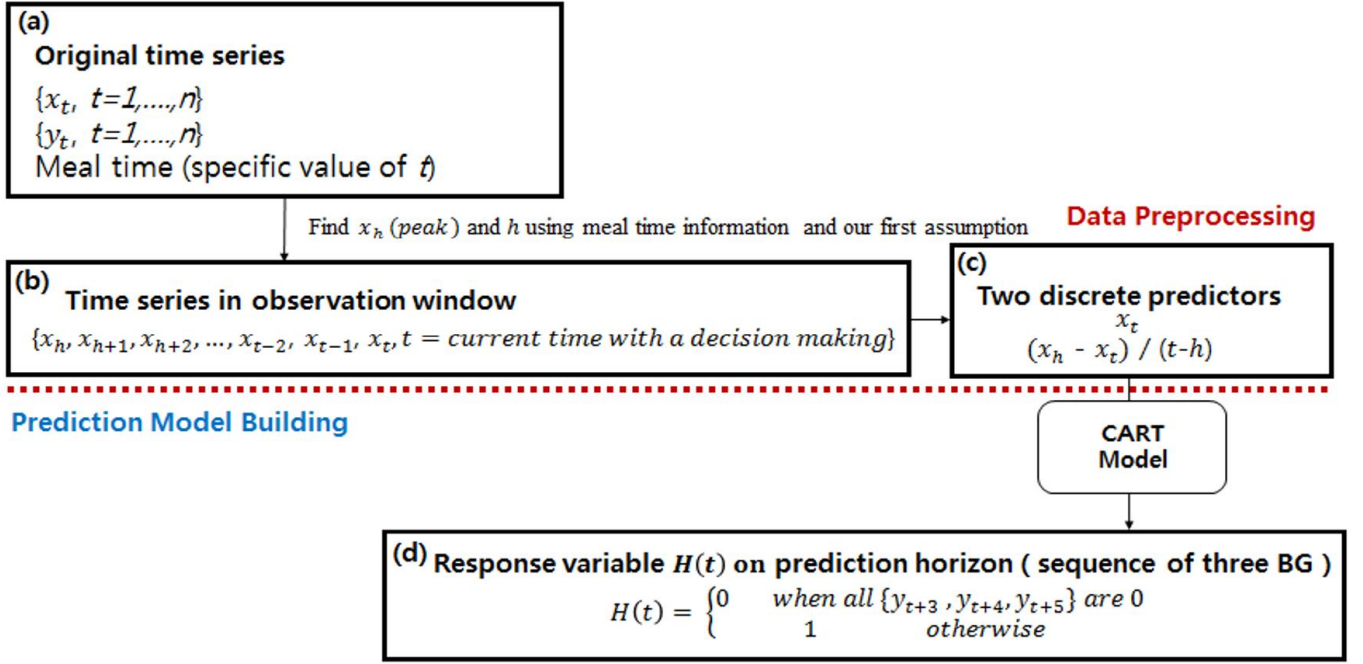

Figure 3. Research phases of our hypoglycemia prediction model

Given that $x_t$ is current BG measurement, the goal is to predict the occurrence of hypoglycemia in the prediction horizon of 15 min { $y_{t+3}, y_{t+4}, y_{t+5}$ } (last box of Figure 3). The first step of preprocessing of original time series is to set a observation window based on the value $x_h$ and $h$. To achieve $x_h$, we set a range of two hours after a meal [36], and found the highest level (peak). Using the starting point $x_h$ and $h$, we set a observation window on the past data (second box of Figure 3). To predict hypoglycemic events within the prediction horizon, previous CGM data { $x_h, x_{h+1}, x_{h+2}, …, x_{t-2}, x_{t-1}, x_t$, h = the time when BG is the highest value after a meal} are used as input for the prediction. With the value of $x_h$, the algorithm will make decisions (predict the occurrence of hypoglycemia within a moving prediction horizon) for every 15 min; so we will make a decision when t=2 hr 15 min after a meal, 2 hr 30 min after a meal, 2 hr 45 min after a meal, 3 hr after a meal, 3 hr 15 min after a meal, 3 hr 30 min after a meal, 3 hr 45 min after a meal, 4 hr after a meal. Each input pattern { $x_h, x_{h+1}, x_{h+2}, …, x_t$ } is formed from a window of the specific length by moving along the CGM time series. We defined response variable of decision tree as the function of hypoglycemia within the prediction horizon, the function $H(t)$, $H(t) = 0$ when all { $y_{t+3}, y_{t+4}, y_{t+5}$} are 0, otherwise 1.

The prediction model used in this study is CART and the input variables are time series features. We didn't use the input set { $x_h, x_{h+1}, x_{h+2}, …, x_{t-2}, x_{t-1}, x_t$ } as it is. Instead of that, we made two predictors (third box of Figure 3) by using only two variables $x_h$ and $x_t$ from the original time series. It improves the efficiency of data storage and computation time when our algorithm will be used in real system. The first predictor is $x_t$ (mmol/L), current BG measurement at a decision point, and $x_t$ is used as it is. The second predictor is the difference between $x_h$ and $x_t$ divided by the time difference $(x_h - x_t) / (t-h)$. The unit of the second predictor is (mmol/L)/min. The positive value of this second predictor indicates a drop in BG. By contrast, the negative value results from the rise in BG. Finally,

$H(t) = D (\{ x_t, (x_h - x_t) / (t-h)\})$
Where a meal time $< h < t$;
the algorithm will make a decision, show H(t), for every 15 min;
D is CART prediction model.

$$H(t) = \begin{cases} 0 & \text{when all } \{y_{t+3}, y_{t+4}, y_{t+5}\} \text{ are } 0 \\ 1 & \text{otherwise} \end{cases} \quad (1)$$

The target attribute $H(t)$ is determined based on the occurrence or non-occurrence of hypoglycemia within a prediction horizon of 15 min. Determination of the length of prediction horizon is also important research decision. Some previous studies set the prediction horizon as a single time stamp, not a specific period like us. We set the prediction horizon as 15-min, not a single time stamp, because hypoglycemia is a symptom lasting for a while. Also, in [40], the duration of hypoglycemia was reported as minimally 15-min. In Figure 4, The sky blue region means the prediction horizon when the proposed algorithm makes current decision at the green circle point. With the course of time, the sky blue region will move to next 15-min time region, and the decision point also will move to the 15-min later time stamp. Table 2 shows a sample of one patient's data after preprocessing. The fifth and sixth columns made by preprocessing, which are used as input predictors and last column is used as the response variable of CART model. The peak $x_h$ is fixed with same meal time, but the current time and BG value $x_t$ changes along the moving PH (Table 3). Based on same meal time, seven decisions of alarming will be made based on the value of two predictors (gray-shaded columns).

**Table 3. Sample of preprocessed data of patient 0**

| Meal Time | Peak Time | Peak Value $x_h$ (mmol/L) | Current Time | Current $x_t$ (mmol/L) | The rate of decrease in BG $(x_h - x_t) / (t-h)$ | Prediction Horizon (PH) | Hypoglycemia in PH |
|---|---|---|---|---|---|---|---|
| 19:07 | 19:32 | 15.7 | 21:07 | 8 | 0.081 mmol/L / min | 21:22 -21:37 | 0 |
| 19:07 | 19:32 | 15.7 | 21:22 | 8.7 | 0.064 mmol/L / min | 21:37 - 21:52 | 0 |
| 19:07 | 19:32 | 15.7 | 21:37 | 8.4 | 0.058 mmol/L / min | 21:52 - 22:07 | 0 |
| … | … | … | … | … | … | … | … |
| 8:42 | 9:17 | 12.7 | 10:42 | 6.6 | 0.072 mmol/L / min | 10:57 - 11:12 | 0 |
| 8:42 | 9:17 | 12.7 | 10:57 | 5.6 | 0.071 mmol/L / min | 11:12 - 11:27 | 0 |
| 8:42 | 9:17 | 12.7 | 11:12 | 4.8 | 0.069 mmol/L / min | 11:27 - 11:42 | 1 |

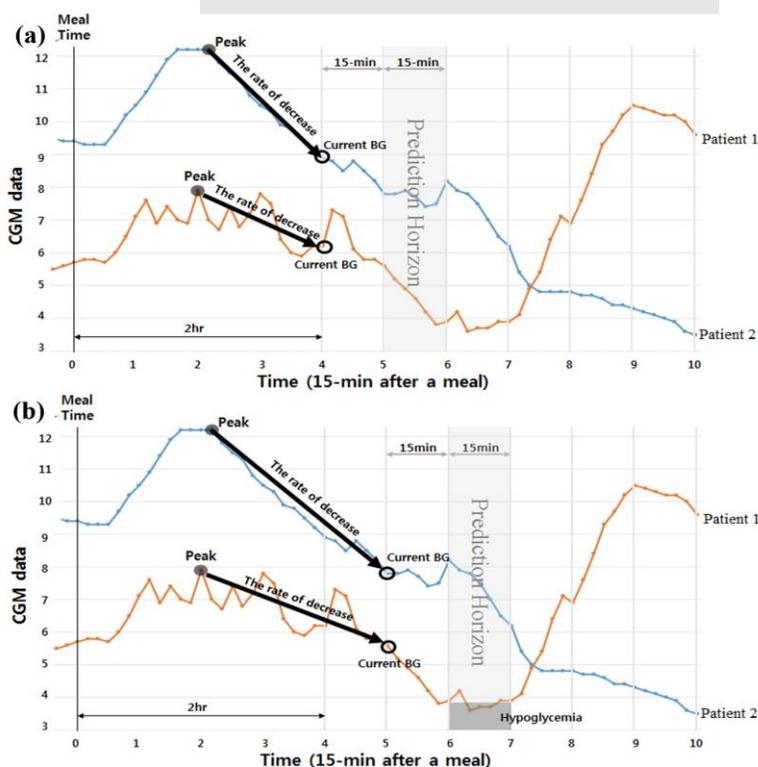

**Figure 4.** Semantic of our prediction model with two patients' data examples; The blue line is BG change of one patient at dinner; the orange line is BG of another patient after breakfast; yellow circle is peak value after a meal; green circle is current BG at the decision.

V. DECISION TREE LEARNING AND PREDICTION

A. *CART Predicion Model with Time Series Features*

CARTs were trained on randomly selected 4/5 of all patients' data, to yield a general model of hypoglycemia prediction regardless of patients' characteristics. If we set our prediction strategy as individual learning model, there was a lack of hypoglycemic events in the training set of each patients. Also, in real situation, patients need to experience hypoglycemia more than 8 to 10 times to let the software learn (find) decision rule (decision tree) from their data - self-contradictorily. The detail process of predicting the hypoglycemia for each events after meals is carried as follows:
1. We applied the preprocessing steps in section *C* to data of each patient, and got the table with two predictors and response variable (like Table 3).
2. Each of the preprocessed dataset was gathered to make one EXCEL file with three columns, two predictors and one response variable. The number of rows of this file was 1867, which is the data from 33 patients for 3-4 days with 2-3 meals per day, and 7 time sections (2hr 15min to 2hr 30min, 2hr 30min to 2hr 45min, …, 3hr 45min to 4hr after a meal) per meal.
3. We divided 1867 rows (data points) into 5 groups with approximately same number of data points (373,373,373,374,374).
4. CART model was generated for four groups' combined data set (4/5 of all data), and remained one group was used as a test set to calculate the corresponding accuracy, sensitivity and specificity of the CART model generated by the 4/5 of data (This method is called 5-fold cross validation). Then, we got 5 CART models with 5 corresponding performances.
5. We did 3 and 4 steps four times; then we got totally 20 CART models and 20 corresponding performances.
6. The best tree with the highest accuracy and sensitivity was tested on each patient's dataset one by one.
7. The best tree with the highest accuracy and sensitivity was tested on each patient groups' dataset.

### B. Misclassficiation Costs and Prunning Level of CART Model

CART model is generated differently depending on misclassification costs. Misclassification costs are the cost happened when a prediction model makes wrong decisions (false positive and false negative). In medical research, the cost of false positive and the cost of false negative are different. For example, if a pre-diagnosis algorithm classifies (diagnosis) that tumor is benign or not, the cost of missing the sign of tumor is higher than the cost of wrong diagnosis of tumor, because pre-diagnosis test is a screening phase before main (expensive) diagnosis test.

For hypoglycemia prediction, the cost of missing hypoglycemia is higher than the cost of false positive. If we miss the sign of hypoglycemia, patients' BG can drop until less than 70 mg/dL within about 30-min. Missed hypoglycemic events can cause more dangerous situation than wrong positive alarm resulting in instant hyperglycemic consequence. However, quantifying the magnitude of risk of both misclassifications is difficult; even we know missed hypoglycemic events are more dangerous, we cannot 'quantify' how much dangerous it is. In the previous study [32], set the cost of false negative as 10 times of the cost of false positive. However, the number 10 is arbitrary. In this study, we set the cost of misclassification based on the academic recommendation of imbalanced data handling.

When a training data set is imbalanced because of less prevalence of hypoglycemia, researchers can modify the misclassification costs to handle the imbalanced effect of each class [41]. In section 3, we explained that Gini's diversity index was used as an impurity measurement index. The Gini's index can be expressed using the misclassification cost. Therefore, each node is determined by the criterion related with misclassification costs; Researchers can reflect both classes on decision rules approximately-equally by using the ratio of the number of objects in each class. It means that we can determine the misclassification costs using the imbalance ratio. As a result, we set the cost of false negative as 15 and the cost of false positive as 1, because the number of hypoglycemic events is one fifteenth of non-hypoglycemic events. The cost matrix of our model is shown in Table 4.

**Table 4. Misclassification costs of our CART model**

|        | Predictive class N | Predictive class H |
|--------|--------------------|--------------------|
| True N | 0                  | 1                  |
| True H | 15                 | 0                  |

Secondly, full-grown tree needs to be pruned, due to its complexity and over-fitting. We test the best pruning level based on cost-complexity measure with 5 fold cross-validation, and the result was depth 3. Also, our clinical advice team said that the tree with depth 3 is most interpretable. Both of cost-complexity measure and expert's knowledge indicated the best depth level of 3. Therefore, we used the depth level of 3 to prune each tree.

### C. Evaluation Method

We used three indices (accuracy, sensitivity and specificity) to evaluate the performance of our prediction method. Especially, sensitivity (true positive ratio) is important index with small number of true class (hypoglycemia). As we mentioned above, we used 5-fold cross-validation with totally 20 iterations. Also, we test our best decision tree among 20 iterations on the data of each patient. In addition, we made patient groups depending on their diabetes types, and test our best model on each groups to find the difference of performance depending on the diabetes types.

## VI. RESULTS

To represent the performance of hypoglycemia prediction, three performance indices were used. They are calculated as follows,

$$Sensitivity = \frac{True\ positive}{True\ positive + False\ negative}$$

$$Specificity = \frac{True\ negative}{True\ negative + False\ positive}$$

$$Accuracy = \frac{True\ positive + True\ negative}{All\ events}$$

where True positive is the number of correctly detected hypoglycemic events 15 min in advance, False negative is the number of undetected impending hypoglycemic events at a decision point, and False positive is the number of wrong predictive decisions that there would be hypoglycemic events within 15 min even though the real BG doesn't fall until 70mg/dL.

*A. 5-fold cross-validation results*

The results are the prediction performance indices of each tree trained on randomly-selected 4/5 of all events, and validated on the other 1/5 of all events. The number of total events is 1867 and 120 events are hypoglycemia. The hypoglycemia complications occurrence rate was approximately 6.4% in our study. Accuracy, sensitivity, and specificity of each tree are shown in Table 5.

Table 5. Accuracy, Sensitivity, and Specificity of each test set; RN means random number allocation

| | **Accuracy (%)** | | | | | **Sensitivity (%)** | | | | | **Specificity (%)** | | | | |
|---|---|---|---|---|---|---|---|---|---|---|---|---|---|---|---|
| | Test 1 | 2 | 3 | 4 | 5 | Test 1 | 2 | 3 | 4 | 5 | Test 1 | 2 | 3 | 4 | 5 |
| **RN.1** | 83 | 79 | 79 | 75 | 80 | 90 | 73 | 70 | 56 | 96 | 83 | 80 | 80 | 76 | 79 |
| **RN.2** | 82 | **86** | 76 | 75 | 80 | 79 | **75** | 85 | 92 | 82 | 82 | **87** | 75 | 73 | 80 |
| **RN.3** | 82 | 86 | 82 | 74 | 80 | 79 | 70 | 77 | 88 | 96 | 82 | 88 | 82 | 73 | 79 |
| **RN.4** | 84 | 80 | 76 | 77 | 80 | 63 | 67 | 80 | 88 | 95 | 86 | 82 | 76 | 76 | 79 |

Because we have already adjusted the imbalance between classes by modifying misclassification costs, the sensitivity is reasonable even with small number of hypoglycemic events. Total 20 replications had accuracy in the range from 74% to 86%, and the average accuracy was 79.8%. The overall (average) sensitivity was 80.05% and the overall specificity was 79.9%.

*B. Individuals prediction performance by the best tree*

We tested our prediction model to the individual data. The performance results of the best tree are summarized in Table 6. The best tree is the tree yielded by the training set except the test set 2 selected by RN.2 (Figure 5).

Table 6. Detailed results of each patient; gray-shaded rows' sensitivity is lower than 50%;
blue-shaded rows' accuracy and specificity is lower than 70%; DM: Diabetes mellitus; DS: Dumping syndrome

| Patient Num. | Total test points | Hypoglycemic points | DM type | Accuracy (%) | Sensitivity (TP ratio, %) | Specificity (TN ratio, %) |
|---|---|---|---|---|---|---|
| 0 | 52 | 3 | 1 | 96 | 100 | 96 |
| 1 | 57 | 5 | 1 | 92 | 100 | 91 |
| 2 | 89 | 5 | 1 | 76 | 20 (1/5) | 80 |
| 3 | 43 | 2 | 1 | 91 | 0 (0/2) | 95 |
| 4 | 80 | 7 | DS | 94 | 100 | 93 |
| 5 | 40 | 4 | 1 | 80 | 100 | 78 |
| 6 | 33 | 4 | 1 | 67 | 100 | 62 |
| 7 | 81 | 3 | 1 | 81 | 67 | 82 |
| 8 | 64 | 4 | 2 | 89 | 75 | 90 |
| 9 | 44 | 2 | 1 | 86 | 100 | 86 |
| 10 | 42 | 2 | 2 | 86 | 50 (1/2) | 88 |
| 11 | 46 | 3 | 1 | 87 | 67 | 88 |
| 12 | 57 | 2 | 1 | 88 | 50 (1/2) | 89 |
| 13 | 56 | 2 | 2 | 100 | 100 | 100 |
| 14 | 49 | 5 | 2 | 73 | 80 | 73 |
| 15 | 56 | 3 | 1 | 84 | 67 | 85 |
| 16 | 43 | 5 | 2 | 81 | 75 | 82 |
| 17 | 69 | 6 | 1 | 86 | 67 | 87 |
| 18 | 40 | 4 | 2 | 100 | 100 | 100 |
| 19 | 40 | 3 | 2 | 80 | 67 | 81 |
| 20 | 41 | 2 | 2 | 83 | 100 | 82 |
| 21 | 68 | 3 | 1 | 67 | 100 | 65 |
| 22 | 66 | 6 | 2 | 80 | 100 | 78 |
| 23 | 59 | 2 | 1 | 81 | 100 | 81 |
| 24 | 54 | 3 | 1 | 83 | 100 | 82 |
| 25 | 65 | 3 | 1 | 85 | 100 | 84 |
| 26 | 43 | 3 | 1 | 77 | 100 | 75 |

| 27 | 65 | 5 | DS | 88 | 80 | 88 |
| 28 | 85 | 6 | 2 | 81 | 83 | 81 |
| 29 | 78 | 2 | 1 | 92 | 100 | 92 |
| 30 | 42 | 7 | 1 | 79 | 86 | 77 |
| 31 | 66 | 6 | 1 | 71 | 100 | 68 |
| **32** | 53 | 2 | 1 | 85 | 0 | 88 |

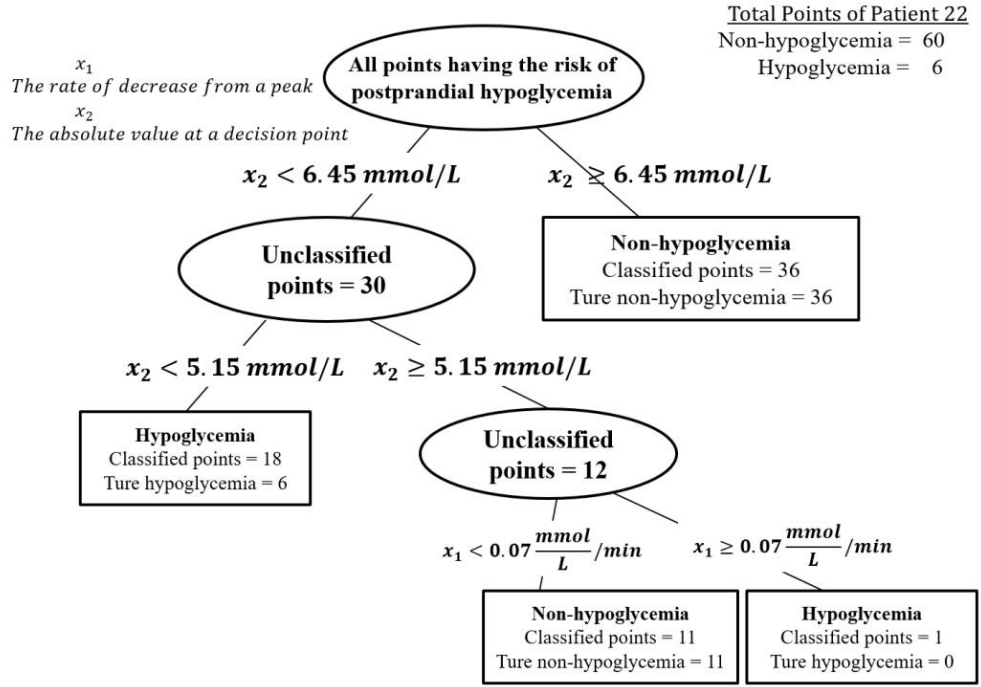

**Figure 5. The best tree of our prediction method with the training set except test set 2 selected by RN.2**

Table 5 shows the individual results on hypoglycemia prediction, in terms of accuracy, sensitivity, and specificity. Except patients 2,3,10,12 and 32, the sensitivity is higher than 50%. Also, except patients 6 and 21, the accuracy and specificity are higher than 70%. Were the model to be used in practice, most patients can prevent hypoglycemic events. While we are still working to improve sensitivity, these results provide proof of concept that the model could alert patients to impending hypoglycemic events.

*C. Individuals Prediction Performances by the Best Tree*

We tested our prediction model to the individual. The best tree is the tree yielded by the training set except the test set 2 selected by the second random number allocation (Fig. 5). Table VI shows the individual results on hypoglycemia prediction, in terms of accuracy, sensitivity, and specificity. Except patients 2, 3, 10, 12 and 31 (the lower 15% of sensitivity), the sensitivity is greater than or equal to 67% for the other 85% of all patients. Also, except patients 6, 21 and 30 (the lower 10% of specificity and accuracy), the accuracy and specificity are greater than or equal to 73%.

Table VII shows the result of our subgroup analysis. The subgroup consists of the sixteen patients who had missed hypoglycemic events. The objective of this subgroup analysis is to examine the lowest glucose level with false negative.

The missed hypoglycemic events per person were mostly one or two. The lowest glucose levels with each false negative alarm are shown in Table VII. The total number of hypoglycemic events is 120, and the number of missed hypoglycemic events among them is 22. Among 22 missed events, only 5 events were severe hypoglycemia, ≤ 2.8 mmol/L. Even though our prediction tree fails to predict some impending hypoglycemic events, 17 events among these 22 missed events (77% of false negative events) were not severe hypoglycemia. Also, almost a half (10/22) of missed events is near to our threshold (from 3.7 to 3.9 mmol/L).

TABLE VI
Detailed results of each patient; Gray-shaded rows are the lower 15% of sensitivity;
Highlighted-Border rows are the lower 10% of specificity and accuracy;
*The patient 25 was participated in this study twice at an interval of two years

| Patient Num. | Total test points | Hypoglycemic points | DM type | Accuracy (%) | Sensitivity (TP ratio, %) | Specificity (TN ratio, %) |
|---|---|---|---|---|---|---|
| 0 | 52 | 3 | 1 | 96 | 100 | 96 |
| 1 | 57 | 2 | 1 | 92 | 100 | 91 |
| 2 | 89 | 5 | 1 | 76 | 20 (1/5) | 80 |
| 3 | 43 | 2 | 1 | 91 | 0 (0/2) | 95 |
| 4 | 80 | 7 | alimentary hypoglycemia | 94 | 100 | 93 |
| 5 | 40 | 4 | 1 | 80 | 100 | 78 |
| 6 | 33 | 4 | 1 | 67 | 100 | 62 |
| 7 | 81 | 3 | 1 | 81 | 67 | 82 |
| 8 | 64 | 4 | 2 | 89 | 75 | 90 |
| 9 | 44 | 2 | 1 | 86 | 100 | 86 |
| 10 | 42 | 2 | 2 | 86 | 50 (1/2) | 88 |
| 11 | 46 | 3 | 1 | 87 | 67 | 88 |
| 12 | 57 | 2 | 1 | 88 | 50 (1/2) | 89 |
| 13 | 56 | 2 | 2 | 100 | 100 | 100 |
| 14 | 49 | 5 | 2 | 73 | 80 | 73 |
| 15 | 56 | 3 | 1 | 84 | 67 | 85 |
| 16 | 43 | 4 | 2 | 81 | 75 | 82 |
| 17 | 69 | 6 | 1 | 86 | 67 | 87 |
| 18 | 40 | 4 | 2 | 100 | 100 | 100 |
| 19 | 40 | 3 | 1 | 80 | 67 | 81 |
| 20 | 41 | 2 | 2 | 83 | 100 | 82 |
| 21 | 68 | 3 | 1 | 67 | 100 | 65 |
| 22 | 66 | 6 | 2 | 80 | 100 | 78 |
| 23 | 59 | 2 | 1 | 81 | 100 | 81 |
| 24 | 54 | 3 | 1 | 83 | 100 | 82 |
| 25(2013)* | 65 | 3 | 1 | 85 | 100 | 84 |
| 25(2015)* | 43 | 3 | 1 | 77 | 100 | 75 |
| 26 | 65 | 5 | alimentary hypoglycemia | 88 | 80 | 88 |
| 27 | 85 | 6 | 2 | 81 | 83 | 81 |
| 28 | 78 | 2 | 1 | 92 | 100 | 92 |
| 29 | 42 | 7 | 1 | 79 | 86 | 77 |
| 30 | 66 | 6 | 1 | 71 | 100 | 68 |
| 31 | 53 | 2 | 1 | 85 | 0(0/2) | 88 |

TABLE VII
The lowest glucose with false negative alarms

| Patient Num. | Sensitivity (%) | Predicted hypoglycemic events | Missed hypoglycemic events | The lowest glucose in PH with false negative (mmol/L) |
|---|---|---|---|---|
| 2 | 20 | 1 | 4 | 3.0, 3.8, 3.7, **2.8** |
| 3 | 0 | 0 | 2 | 3.1, **2.2** |
| 7 | 67 | 2 | 1 | 3.7 |
| 8 | 75 | 3 | 1 | 3.4 |
| 10 | 50 | 1 | 1 | 3.8 |
| 11 | 67 | 2 | 1 | 3.8 |
| 12 | 50 | 1 | 1 | 2.9 |
| 14 | 80 | 4 | 1 | 3.8 |
| 15 | 67 | 2 | 1 | 3.8 |
| 16 | 75 | 3 | 1 | **2.7** |
| 17 | 67 | 4 | 2 | 3.9, **2.4** |
| 19 | 67 | 2 | 1 | 3.7 |
| 26 | 80 | 4 | 1 | 3.0 |
| 27 | 83 | 5 | 1 | 3.9 |
| 29 | 86 | 6 | 1 | 3.4 |
| 31 | 0 | 0 | 2 | **2.6**, 3.2 |

## D. Comparison of prediction results among patient groups

When we compared the prediction performance shown in Table VI with two patient groups ( type 1 diabetes group, n = 21; type 2 diabetes group, n = 9) by one-way ANOVA using MATLAB, there was no significant difference in group mean of sensitivity between type 1 and type 2 diabetes groups with $\alpha = 0.05$ (p-value > 0.05). The sensitivity (p-value > 0.05) and specificity (p-value > 0.05) of two groups are also significantly not different with $\alpha = 0.05$.

## VII. DISCUSSION AND CONCLUSION

We have described a machine learning approach to predict hypoglycemia and presented the results of our recent evaluations. Although daytime hypoglycemia raises several complications like car accidents, and causes psychological burden carrying less use of insulin injection, study of prediction of daytime hypoglycemia was not enough. In this study, we focus on daytime hypoglycemia prediction. For predicting the risk of hypoglycemia, we used CART analysis. With this approach, no complicated models or algorithms need to be considered. As a result of 5-fold cross-validation, the overall accuracy, specificity and sensitivity were impressive even though we used only CGM data with 15-min *L*, also, we had enough subjects for both diabetes types. The CART model is known to be sensitive to training set, because the structure of tree varies depending on selected training set. Even though there was this issue of tree, the range of accuracy and sensitivity wasn't large, which means the pruned tree structures are similar among different test sets, and our prediction model is enough generic, regardless of patient. As we mentioned above, our model predicts onset events regardless of patients since the blood glucose variations over time have similar properties in any diabetic patient. This assumption was proved by the comparison result of one-way ANOVA in subsection C. In addition to addressing generic event-based prediction model, we also evaluated our best tree on each patient' data set. Daytime hypoglycemia is more challenging to predict, but the result in Table 6 looks encouraging. Except a few patients, our method predicts each event of each patient accurately. Our proposed model has enough significance to inform reliably predicted risk, and has two practical perspectives; guaranteed time lag and simple input data (CGM data and simple announcement of meals). Our models can be used in systems to give early warnings of potential hypoglycemia sufficiently early that a user can increase her BG level in time to avert it.

Future studies may seek to identify new predictors to make the *L* much longer than current model, and to improve prediction accuracy. In addition, further study can be conducted on prediction with other machine learning methods.

## VIII. ACKNOWLEDGEMENT

This research was supported by the MSIP (Ministry of Science, ICT and Future Planning), Korea, under the "ICT Consilience Creative Program" (IITP-2015-R0346-15-1007) supervised by the IITP (Institute for Information & communications Technology Promotion)